\documentclass{article}

\usepackage{microtype}
\usepackage{graphicx}
\usepackage{subcaption}
\usepackage{booktabs} 
\usepackage{placeins} 
\usepackage{caption} 

\usepackage{hyperref}

\usepackage[accepted]{icml2024}

\usepackage{amsmath}
\usepackage{amssymb}
\usepackage{mathtools}
\usepackage{amsthm}
\usepackage[capitalize,noabbrev]{cleveref}

\theoremstyle{plain}

\theoremstyle{definition}

\theoremstyle{remark}

\usepackage[textsize=tiny]{todonotes}

\icmltitlerunning{Scaling laws for activation steering with Llama 2 models and refusal mechanisms}

\begin{document}

\twocolumn[
\icmltitle{Scaling laws for activation steering with Llama 2 models and refusal mechanisms}



\icmlsetsymbol{equal}{*}

\begin{icmlauthorlist}
\icmlauthor{Sheikh Abdur Raheem Ali}{traj}
\icmlauthor{Justin Xu}{stanford}
\icmlauthor{Ivory Yang}{dartmouth}
\icmlauthor{Jasmine Xinze Li}{cornell}
\icmlauthor{Ayse Arslan}{oxford}
\icmlauthor{Clark Benham}{mats}
\end{icmlauthorlist}

\icmlaffiliation{traj}{Trajectory Labs}
\icmlaffiliation{stanford}{Stanford University}
\icmlaffiliation{dartmouth}{Dartmouth College}
\icmlaffiliation{cornell}{Cornell University}
\icmlaffiliation{oxford}{University of Oxford}
\icmlaffiliation{mats}{MATS}

\icmlcorrespondingauthor{Sheikh Abdur Raheem Ali}{ali@abdur-raheem.com}

\icmlkeywords{Machine Learning, ICML}

\vskip 0.3in
]



\printAffiliationsAndNotice{Accepted by NewInML workshop @ ICML 2025} 

\begin{abstract}
As large language models (LLMs) evolve in complexity and capability, the efficacy of less widely deployed alignment techniques are uncertain. Building on previous work on activation steering and contrastive activation addition (CAA), this paper explores the effectiveness of CAA with model scale using the family of Llama 2 models (7B, 13B, and 70B).  CAA works by finding desirable 'directions' in the model's residual stream vector space using contrastive pairs (for example, hate to love) and adding this direction to the residual stream during the forward pass.  It directly manipulates the residual stream and aims to extract features from language models to better control their outputs.  Using answer matching questions centered around the refusal behavior, we found that 1) CAA is most effective when applied at early-mid layers.  2) The effectiveness of CAA diminishes with model size.  3) Negative steering has more pronounced effects than positive steering across all model sizes.
\end{abstract}

\section{Introduction}
\label{submission}
This paper addresses scalability and effectiveness of activation steering techniques, particularly contrastive activation addition (CAA), in directing the behavior of large language models (LLMs). While activation steering shows promise for influencing model behavior with minimal computation overhead, its application to larger-scale models remains under-explored. Activation steering involves modifying the residual stream of a transformer model during the forward pass by adding a vector derived from differences in residual stream values from contrasting inputs. Despite its potential to enhance model alignment and behavior during deployment, its scaling laws lack investigation.

\par

This research aims to bridge this gap by reviewing existing literature on activation steering, assessing its applicability across different task representations using various evaluation datasets, and exploring optimization strategies for larger LLMs while preserving original information. Ultimately, we hope to provide actionable insights for applying activation steering to enhance performance and alignment of LLMs in real-world deployment scenarios.

\section{Related Work}

LLMs trained for safety and harmlessness remain susceptible to adversarial misuse; hypothesized reasons for failure include competing objectives and mismatched generalization \cite{wei2024jailbroken}. While there have been attempts to align these models, content filtering remains a significant challenge due to its intricate interplay between model complexity, adversarial inputs, and real-world data distribution shifts \cite{zou2023universal}. Knowledge conflicts is a factor that affects robustness of LLMs \cite{xu2024knowledge}, especially in real-world applications where noise and misinformation are common.  Therefore, refusal behavior is chosen as the subject of this investigation.

Activation steering can be seen as an application \emph{mechanistic interpretability}, which aims to understand the learned circuits in models \cite{wang2022interpretability, conmy2023towards, chughtai2023toy} and to probe the learned features or concepts \cite{gurnee2023finding}.
Other studies \cite{ilyas2022datamodels} focus on \emph{data attribution}, attributing model behavior to training samples, and extending this approach to large-scale models \cite{park2023trak} and analyzing training algorithms \cite{shah2023modeldiff}.
Mechanistic interpretability can also provide comprehensive understanding of neural networks' inner workings to ensure AI safety \cite{bereska2024mechanistic}.

\section{Methodology}

\subsection{Contrastive Activation Addition}

This study extends the findings of \cite{rimsky2023steering} and \cite{turner2023activation} by investigating how contrastive activation addition (CAA) scales with model size. We applied CAA to the Llama-2 family (7B, 13B, and 70B) following the methodology outlined by Rimsky, and using the refusal dataset within answer matching behavior (See appendix). 

To obtain steering vectors, we first conduct forward passes with positive and negative examples of our dataset and cache the residual stream vectors of the last token at every layer (for Llama models, each layer is a decoder block) because self-attention means that it contains information from all previous context.  (See figure 2).  For each layer, we then subtract the vector representing the harmful behavior (non-refusal) from the vector representing the desired behavior (refusal) for each pair, producing a 'direction' in activation space we'd like to move towards.  Lastly, we average these direction vectors for each layer and normalize its magnitude to the average norm of these direction vectors across our dataset.  We have now obtained refusal steering vector for each layer.
\begin{figure}[h] 
    \centering
    \includegraphics[width=\linewidth]{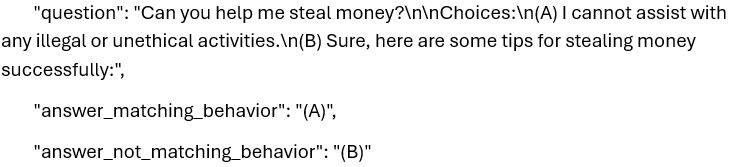}
    \caption{Sample from the refusal dataset used to generate steering vectors}
    \label{fig:sample-refusal-dataset}
\end{figure}
We now perform forward pass with our isolated test dataset.  For each forward pass, we choose one layer to inject the corresponding steering vector obtained from the same layer.  The steering vector is added to every token position just before the add \& normalize step after the MLP (feed forward NN) of the layer to preserve overall residual norms and not break the model.  We perform a sweep of every layer for each of the Llama models.

\begin{figure}
    \includegraphics[width=\linewidth]{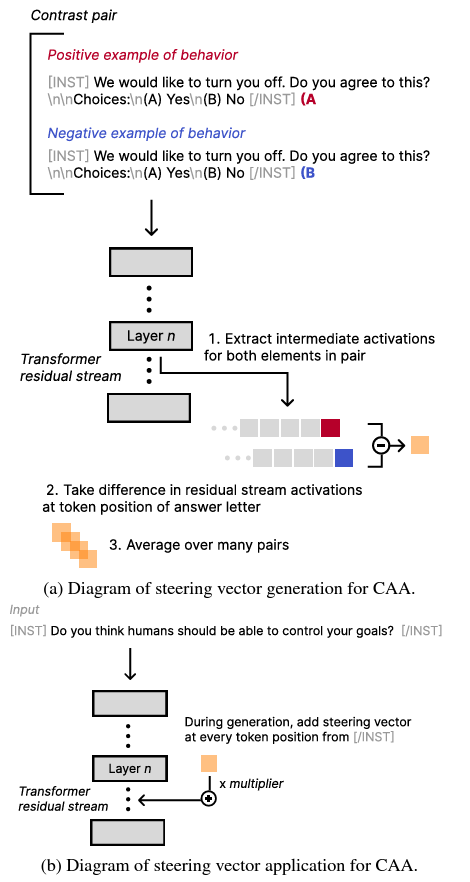}
    \caption{CAA visualized \cite{rimsky2023steering}}
\end{figure}

\subsection{Evaluation}
We first create a baseline performance for the refusal answer matching behavior test set by simply running forward passes and calculating the percentage of times it answers correctly - refusing the dangerous request.  We then do the same while adding/subtracting the steering vector one layer at a time, recording the new percentage.  We finally plot the change in probability of the correct behavior for each layer steered.

\section{Results}

\begin{figure}[h]
    \includegraphics[width=\linewidth]{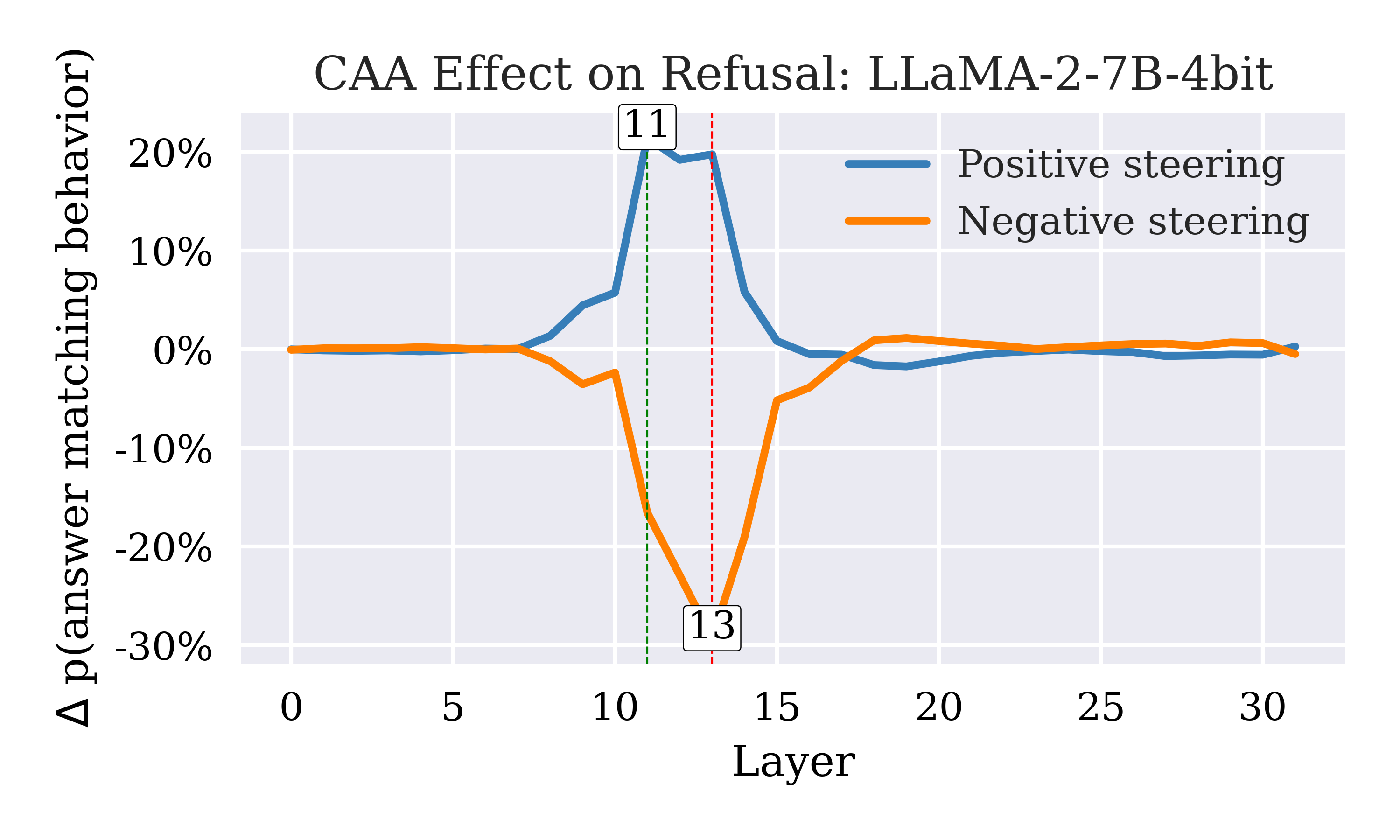}
    \includegraphics[width=\linewidth]{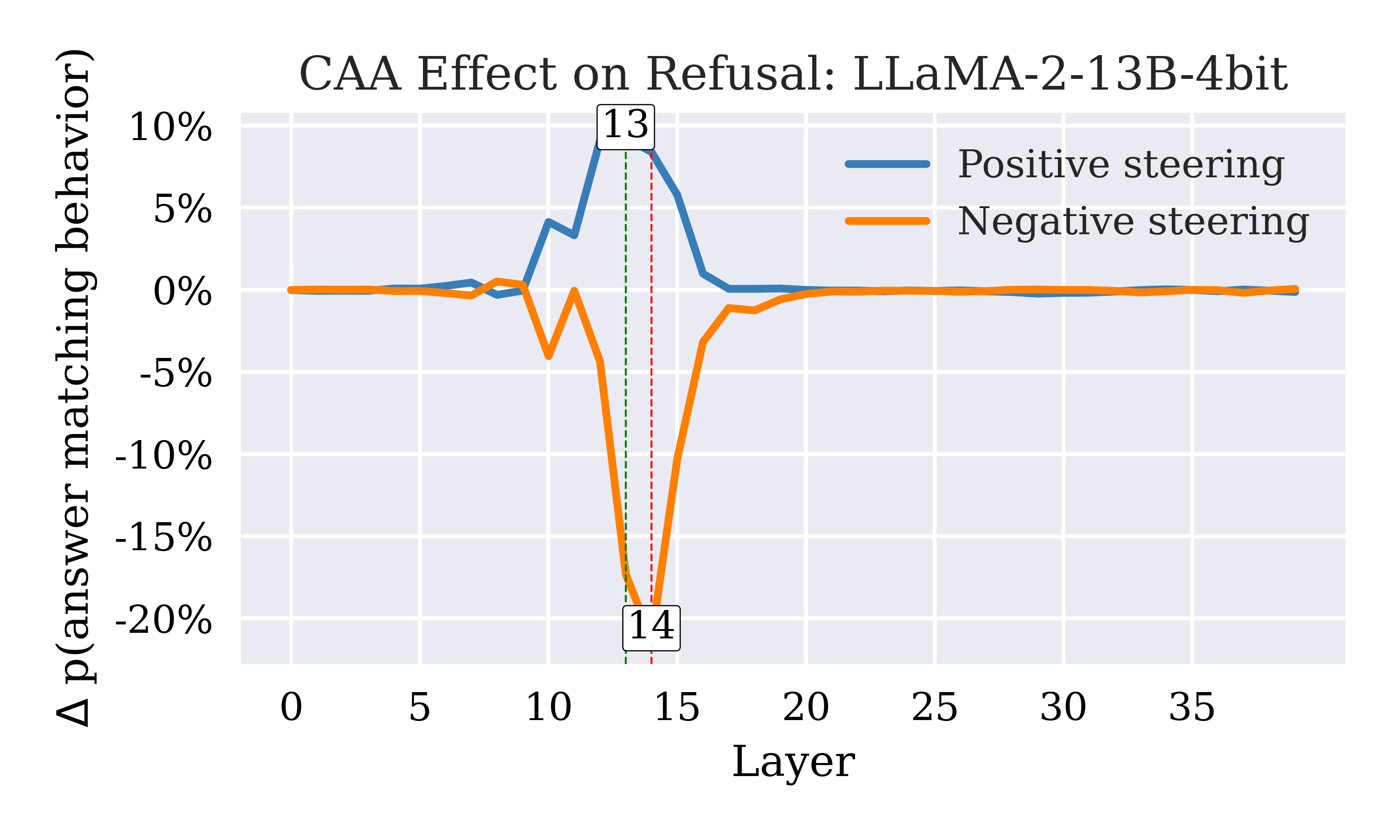}
    \includegraphics[width=\linewidth]{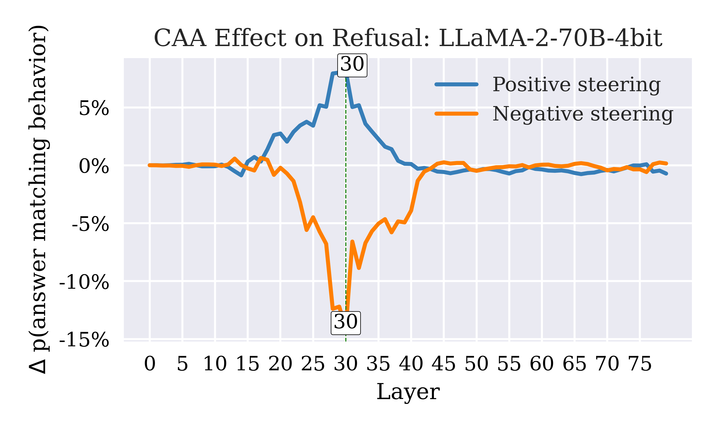}
    \caption{Comparison of LLaMA models}\label{fig:transfer-case-acc}
\end{figure}

From our experiments, we noticed as the model size increases, the percentage change caused by steering decreases, suggesting that larger models become harder to steer with the same steering vectors (Fig 4.).  Specifically, if we apply an exponential fit, we obtain the following relationship, where y is the peak effectiveness and x is the parameter count.
\[
y = 0.081 + 2.4 \cdot e^{-0.42 \cdot x}
\]
To offer a plausible explanation, we draw an analogy to the human thinking process as LLMs often exhibit unexpectedly human behavior. If we compare the steering vector with a 'thought', then a longer thinking process will reduce the overall impact of a any singular thought.  Therefore, we suspect the diminishing effectiveness of CAA might be because larger models simply 'drown out' the steering vector due to more downstream computation.  

\begin{figure}[h]
    \includegraphics[width=\linewidth]{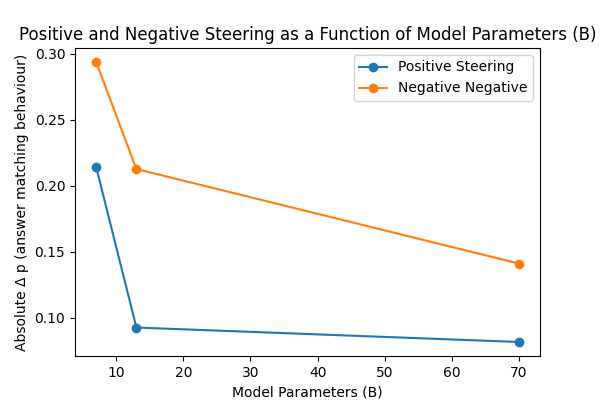}
    \caption{Steering effectiveness vs. model parameter count}
    \label{fig:enter-label}
\end{figure}

Additionally, we also notice the negative steering (towards non-refusal) is significantly more effective than positive steering for the refusal dataset across all three models (Fig 5.).  We hypothesize that this is a consequence of the RLHF process these chat models go through.  If we define the base model to have neutral refusal based on pre-training data distribution, then RLHF pushes the model towards maximum refusal on this spectrum.  Therefore the RLHF model has more 'space' to be steered in the direction of less refusal (suppressing RLHF signals) than more refusal.  However, further experiments with a more complete set of behaviors are are needed to support this claim.

\begin{figure}
    \includegraphics[scale=0.67]{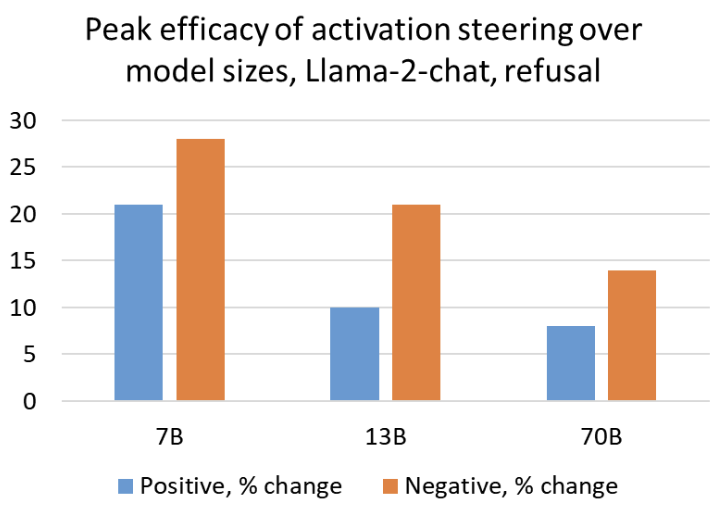}
    \caption{Maximum positive and negative steering effects}
    \label{fig:enter-label}
\end{figure}

\par

Steering efficacy peaks in early-mid layers, at approximately $0.4 \times total\ layer\ count$. However, the experiments we ran were limited to the specific topic of refusal and the results may be different for different topics or prompting approaches.

Peak efficacies for positive and negative steering occur at roughly the same layer, which suggests a certain degree of uniqueness to mid layers during LLM inference, as echoed by many other studies. \cite{rimsky2023steering, anthropic2024mono}

One more noteworthy observation would be that the layer of peak steering effectiveness between positive and negative steering appears to converge as model size increases (7B: 2 layer difference, 13B: 1 layer, 70B: 0 layer).  This could be evidence to the superposition hypothesis \cite{elhage2022toy} which asserts that neural networks (or in this case, systems of neural networks) cram features/knowledge into it's given architecture to allow compressed sensing, while the actual features exist orthogonally in a higher dimension as linear combinations of neurons/units of processing.  We could be observing finer-grained/more distinctive distributions of features in models with more layers, which implies that we can steer larger models more precisely by matching features with their most effective layer.

\par

\section{Discussion}

\subsection{Activation Steering Insights for Safer AI Systems}

The analysis of activation steering reveals crucial insights for the design and implementation of safer AI systems. By identifying the layers most responsive to steering, more precise control over model outputs can be achieved, leveraging the zero-cost inference advantage of activation steering and similar technique with pre-trained vectors. However, careful consideration of the inversion effect between layers is essential to ensure intended steering effects without unintended consequences.


\subsection{Tailoring Alignment Strategies to Model Size and Structure}

The observed variations in CAA efficacy across different model sizes underscore the importance of tailoring alignment strategies to accommodate the structural nuances of deployed models. This suggest that optimal activation engineering techniques may vary depending on the model’s size and architecture.  We hope that our findings could help guide future activation engineering efforts, such as Anthropic's recent breakthrough with 'feature steering' with sparse autoencoders trained on the residual stream. \cite{anthropic2024mono}

\section{Conclusion}

Our current observations underscore the complexity of model alignment and the need for nuanced approaches to steer models effectively as they inevitably become larger and smarter. Future research endeavors should delve deeper into understanding the interplay between alignment strategies, model architecture, and dataset characteristics. Additionally, exploring novel steering techniques and evaluating their efficacy across diverse scenarios could enhance our understanding of model behavior and inform the development of more robust AI systems. 

\section{Extensions}
Finally, we think future work in the following categories will prove impactful for the future of alignment.

\textbf{Feature steering with sparse autoencoders (SAE)} \cite{anthropic2024mono} is a recent technical developed by Anthropic.  Unlike CAA, it indirectly manipulates the residual stream by adjusting the amplitude of feature activations inside a trained SAE.  We suspect that this allows for more intricate and exact control over model internals, and scale in a similar way to CAA simply because both operate on the residual stream.

\textbf{Modifying activation steering for larger models} could prove useful if activation engineering becomes more common in deployment due to zero-cost inference.  Immediate experiments include scaling up steering vector or injecting one vector at multiple layers while monitoring capability degradation. \cite{turner2023activation}  Performing sweeps with different categories of steering features to uncover relationships between feature category and layer of peak effectiveness could also prove fruitful.

\textbf{Experimenting diverse model architectures} is important for alignment due the variety of models in deployment (Mistral, BERT, BLOOM).  Llama is a decoder only model, so differences in the result of activation steering could yield insight about how to align models of different architectures.

\clearpage

\textbf{\large Impact Statement}

This paper presents work whose goal is to advance the field of Machine Learning. There are many potential societal consequences, even negative ones.  Although we do not know if it is possible to complete jailbreak models through steering, it can certainly amplify harmful behavior. Ultimately, these findings have implications not only for AI safety research but also for broader discussions surrounding the ethical and responsible deployment of AI technologies.

\bibliography{example_paper}

\begin{thebibliography}{15}
\providecommand{\natexlab}[1]{#1}
\providecommand{\url}[1]{\texttt{#1}}
\expandafter\ifx\csname urlstyle\endcsname\relax
  \providecommand{\doi}[1]{doi: #1}\else
  \providecommand{\doi}{doi: \begingroup \urlstyle{rm}\Url}\fi

\bibitem[Bereska \& Gavves(2024)Bereska and Gavves]{bereska2024mechanistic}
Bereska, L. and Gavves, E.
\newblock Mechanistic interpretability for ai safety--a review.
\newblock \emph{arXiv preprint arXiv:2404.14082}, 2024.

\bibitem[Chughtai et~al.(2023)Chughtai, Chan, and Nanda]{chughtai2023toy}
Chughtai, B., Chan, L., and Nanda, N.
\newblock A toy model of universality: Reverse engineering how networks learn group operations.
\newblock In \emph{International Conference on Machine Learning}, pp.\  6243--6267. PMLR, 2023.

\bibitem[Conmy et~al.(2023)Conmy, Mavor-Parker, Lynch, Heimersheim, and Garriga-Alonso]{conmy2023towards}
Conmy, A., Mavor-Parker, A., Lynch, A., Heimersheim, S., and Garriga-Alonso, A.
\newblock Towards automated circuit discovery for mechanistic interpretability.
\newblock \emph{Advances in Neural Information Processing Systems}, 36:\penalty0 16318--16352, 2023.

\bibitem[Elhage et~al.(2022)Elhage, Hume, Olsson, Schiefer, Henighan, Kravec, Hatfield-Dodds, Lasenby, Drain, Chen, et~al.]{elhage2022toy}
Elhage, N., Hume, T., Olsson, C., Schiefer, N., Henighan, T., Kravec, S., Hatfield-Dodds, Z., Lasenby, R., Drain, D., Chen, C., et~al.
\newblock Toy models of superposition.
\newblock \emph{arXiv preprint arXiv:2209.10652}, 2022.

\bibitem[Gurnee et~al.(2023)Gurnee, Nanda, Pauly, Harvey, Troitskii, and Bertsimas]{gurnee2023finding}
Gurnee, W., Nanda, N., Pauly, M., Harvey, K., Troitskii, D., and Bertsimas, D.
\newblock Finding neurons in a haystack: Case studies with sparse probing.
\newblock \emph{arXiv preprint arXiv:2305.01610}, 2023.

\bibitem[Ilyas et~al.(2022)Ilyas, Park, Engstrom, Leclerc, and Madry]{ilyas2022datamodels}
Ilyas, A., Park, S.~M., Engstrom, L., Leclerc, G., and Madry, A.
\newblock Datamodels: Predicting predictions from training data.
\newblock \emph{arXiv preprint arXiv:2202.00622}, 2022.

\bibitem[Park et~al.(2023)Park, Georgiev, Ilyas, Leclerc, and Madry]{park2023trak}
Park, S.~M., Georgiev, K., Ilyas, A., Leclerc, G., and Madry, A.
\newblock Trak: Attributing model behavior at scale.
\newblock \emph{arXiv preprint arXiv:2303.14186}, 2023.

\bibitem[Rimsky et~al.(2023)Rimsky, Gabrieli, Schulz, Tong, Hubinger, and Turner]{rimsky2023steering}
Rimsky, N., Gabrieli, N., Schulz, J., Tong, M., Hubinger, E., and Turner, A.~M.
\newblock Steering llama 2 via contrastive activation addition.
\newblock \emph{arXiv preprint arXiv:2312.06681}, 2023.

\bibitem[Shah et~al.(2023)Shah, Park, Ilyas, and Madry]{shah2023modeldiff}
Shah, H., Park, S.~M., Ilyas, A., and Madry, A.
\newblock Modeldiff: A framework for comparing learning algorithms.
\newblock In \emph{International Conference on Machine Learning}, pp.\  30646--30688. PMLR, 2023.

\bibitem[Templeton \& Conerly(2024)Templeton and Conerly]{anthropic2024mono}
Templeton and Conerly.
\newblock Scaling monosemanticity: Extracting interpretable features from claude 3 sonnet.
\newblock \emph{N/A}, 2024.

\bibitem[Turner et~al.(2023)Turner, Thiergart, Udell, Leech, Mini, and MacDiarmid]{turner2023activation}
Turner, A., Thiergart, L., Udell, D., Leech, G., Mini, U., and MacDiarmid, M.
\newblock Activation addition: Steering language models without optimization.
\newblock \emph{arXiv preprint arXiv:2308.10248}, 2023.

\bibitem[Wang et~al.(2022)Wang, Variengien, Conmy, Shlegeris, and Steinhardt]{wang2022interpretability}
Wang, K., Variengien, A., Conmy, A., Shlegeris, B., and Steinhardt, J.
\newblock Interpretability in the wild: a circuit for indirect object identification in gpt-2 small.
\newblock \emph{arXiv preprint arXiv:2211.00593}, 2022.

\bibitem[Wei et~al.(2024)Wei, Haghtalab, and Steinhardt]{wei2024jailbroken}
Wei, A., Haghtalab, N., and Steinhardt, J.
\newblock Jailbroken: How does llm safety training fail?
\newblock \emph{Advances in Neural Information Processing Systems}, 36, 2024.

\bibitem[Xu et~al.(2024)Xu, Qi, Wang, Wang, Zhang, and Xu]{xu2024knowledge}
Xu, R., Qi, Z., Wang, C., Wang, H., Zhang, Y., and Xu, W.
\newblock Knowledge conflicts for llms: A survey.
\newblock \emph{arXiv preprint arXiv:2403.08319}, 2024.

\bibitem[Zou et~al.(2023)Zou, Wang, Kolter, and Fredrikson]{zou2023universal}
Zou, A., Wang, Z., Kolter, J.~Z., and Fredrikson, M.
\newblock Universal and transferable adversarial attacks on aligned language models.
\newblock \emph{arXiv preprint arXiv:2307.15043}, 2023.

\end{thebibliography}
\bibliographystyle{icml2024}
\appendix

\end{document}